\documentclass[twoside,leqno,twocolumn]{article}

\usepackage[letterpaper]{geometry}

\usepackage{ltexpprt}

\usepackage{amsmath,amssymb,amsfonts}
\usepackage{algorithmic}
\usepackage{graphicx}
\usepackage{textcomp}
\usepackage{xcolor}
\usepackage{float}
\usepackage[ruled]{algorithm}
\usepackage{bm} 
\usepackage{physics} 
\usepackage{graphicx}
\usepackage{subfigure}

\usepackage{appendix}

\begin{document}

\title{\Large Adaptive Transfer Learning of Multi-View Time Series Classification}
\author{Donglin Zhan\thanks{Sichuan University, Equal Contribution}
\and Shiyu Yi\thanks{Sichuan University, Equal Contribution}
\and Dongli Xu\thanks{Harbin Engineering University}
\and Xiao Yu\thanks{University of Electronic Science and Technology of China}
\and Denglin Jiang\thanks{New York University}
\and Siqi Yu\thanks{Sichuan University}
\and Haoting Zhang\thanks{University of California, Berkeley}
\and Wenfang Shangguan\thanks{Sichuan University}
\and Weihua Zhang\thanks{Sichuan University}
}

\date{}

\maketitle







\begin{abstract} \small\baselineskip=9pt 
Time Series Classification (TSC) has been an important and challenging task in data mining, especially on multivariate time series and multi-view time series data sets. Meanwhile, transfer learning has been widely applied in computer vision and natural language processing applications to improve deep neural network’s generalization capabilities.
However, very few previous works applied transfer learning framework to time series mining problems. Particularly, the technique of measuring similarities between source domain and target domain based on dynamic representation such as density estimation with importance sampling has never been combined with transfer learning framework. 
In this paper, we first proposed a general adaptive transfer learning framework for multi-view time series data, which shows strong ability in storing inter-view importance value in the process of knowledge transfer. Next, we represented inter-view importance through some time series similarity measurements and approximated the posterior distribution in latent space for the importance sampling  via density estimation techniques. We then computed the matrix norm of sampled importance value, which controls the degree of knowledge transfer in pre-training process. We further evaluated our work, applied it to many other time series classification tasks, and observed that our architecture maintained desirable generalization ability. Finally, we concluded that our framework could be adapted with deep learning techniques to receive significant model performance improvements. 
\end{abstract}

\section{Introduction}

Over the past few decades, with the ever-accelerated development of machine learning and deep learning, many previous bottleneck problems in finance, healthcare, media, and other application fields could be transformed into time series classification (TSC) problems and then be solved with the help of advanced deep learning tools, such disease diagnosis from time series of physiological parameters, classifying heart arrhythmias from ECG signals \cite{y1}, and human activity recognition \cite{y2}. Among them, Deep Neural Networks (DNNs) such as Long Short Term Memory Recurrent Neural Networks (LSTM-RNNs) \cite{y3} and 1-dimensional convolution neural networks (CNNs) \cite{y4, y5, y6} have achieved state-of-the-art results. However, under circumstances where access to a large labeled training data set is not available, which is always the case with time series data, these fancy DNNs overfit terribly \cite{y7, y8}. For example, even though Convolutional Neural Networks (CNNs) could have achieved impressive model performance figures when combined with the Dynamic Time Warping (DTW) algorithm (1NN-DTW)\cite{y9}, it suffers from the problem of over-fitting once applied in the Time Series Classification (TSC) tasks, and this would become worse when there are not enough samples or when patterns of the data are time-variant \cite{y10}. In short, due to the difficulty in collecting and annotating time-series data, DNNs could hardly be applied to small-scale time series data sets \cite{y11}.

Therefore, studies that focus on solving time series classification (TSC) with other techniques have appeared in the past decade. Although many metrics are proposed in the previous works (e.g., Dynamic Time Wrapping (DTW) \cite{y12}, edit distance \cite{y13}, elastic distance\cite{y14}, they concentrate only on single-view \cite{y15} or univariate time series (u.t.s.) classification tasks \cite{y16,y17} instead of those on multi-view time series. Moreover, since these traditional methods largely count on tremendous sample size and labels, they tend to receive poor performance especially in model efficiency and accuracy.

Thanks to the increasing number of various sensors, information of the same object could be collected from multiple perspectives. Such mutually enriched and supported information could largely help machine learning tasks by offering higher quality, more diverse information, and thus increase model performance\cite{y18}. Compared with traditional single-view methods, multi-view learning yields better results and has received an increasing amount of attention over the past few years \cite{y19, y20}. Recent popular multi-view learning methods include collaborative training, multi-core learning and subspace learning \cite{y19}. Hence, if applicable, multi-view data are usually preferred over single-view data. However, in Time Series Classification (TSC) tasks, either existing methods for time series data classification only focus on single-view data and the benefits of mutual-support multiple views are not taken into account, or existing multi-view learning methods cannot be appropriately applied to multi-view time series, because many of the unique properties of multi-view time series are ignored.

On the other hand, transfer learning received quite an amount of research attention recently. It is a research problem that focuses on storing knowledge gained while solving one problem and applying it to a different but related problem. In other words, transfer learning methods could learn from a source task which has enough labeled data collections \cite{y21}. The method yields rather satisfying model performance when used in computer vision, \cite{y22}, \cite{y23}, social media analytics \cite{y24}–\cite{y25}, anomaly detection \cite{y26}. However, not so much work has examined its performance in time-series classification problems. 

In light of all these challenges, we proposed a novel approach to deal with classification tasks on multi-view time series data sets through transfer learning. Overall, our contributions are as follows:


\begin{itemize}
    \item We proposed a dynamic inter-view importance measurement to capture the correlation based on different views more robustly, which can enhance interpretability when combined with knowledge transfer. 

    \item We combined cutting-edge density estimation techniques with classical univariate and multivariate time series distance measurements. Density estimation tools are used to approximate the posterior distribution based on similarity features captured by time series distance algorithms. 

    \item We proposed a concept of adaptive transfer degree based on
    multi-view time series data, which will be sampled from approximated posterior distribution. During the training process in source domains and views, the proposed transfer degree can control the degree of knowledge transfer. 
    
    \item We validated our framework's performance on some widely used time series classification models and experimented results on several open data sets to show that our proposed method truly generalizes, and could significantly improve the classification accuracy.
\end{itemize}


\begin{figure*}[!ht]
    \centering
    \includegraphics[width=0.8\textwidth]{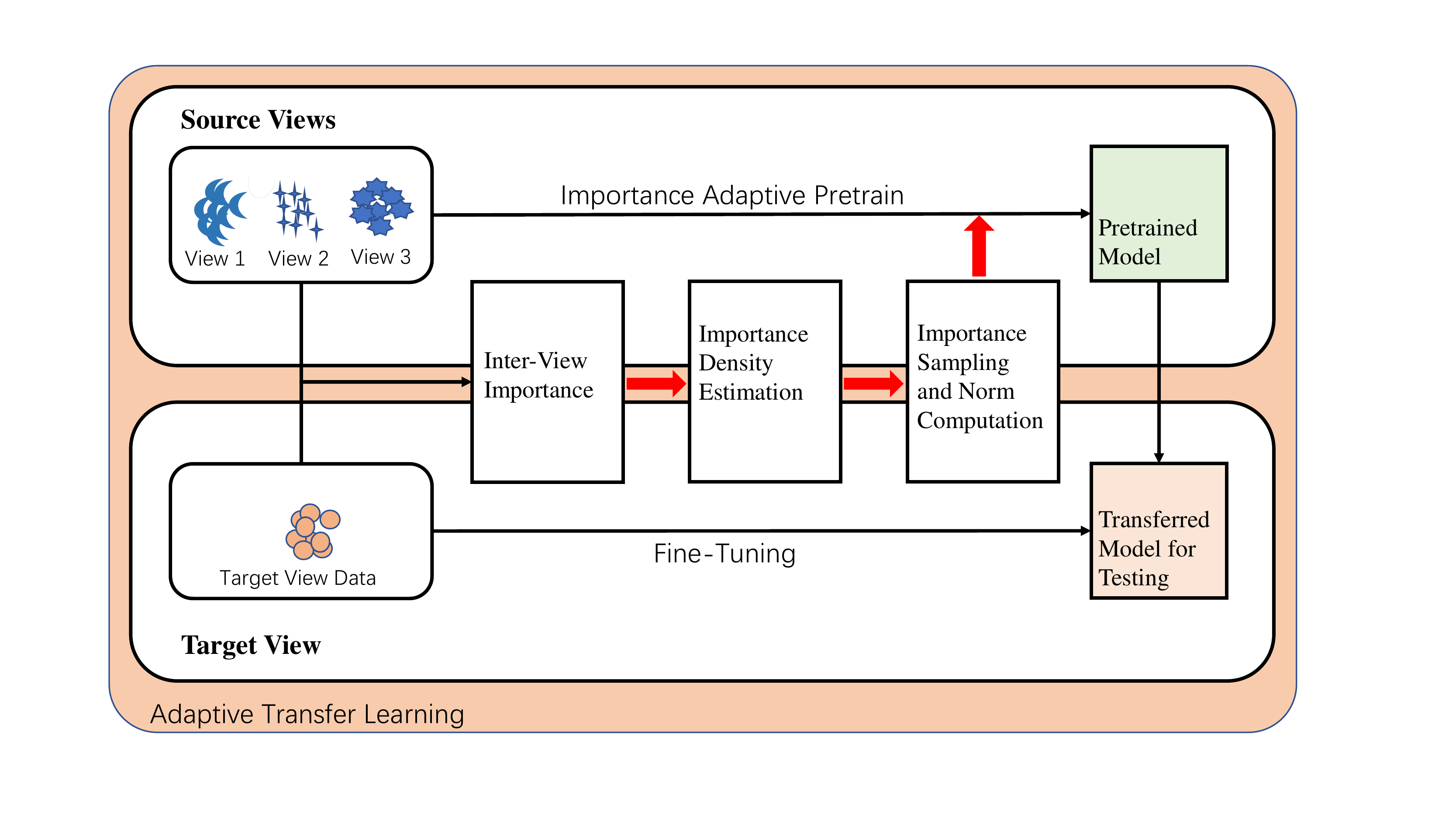}
    \caption{Framework of Adaptive Transfer Learning for Multi-View Time Series Classification }
    \label{Figur.1_fr}
\end{figure*}

\section{Related Work}

Since deep learning models exhibited impressive performance in many application fields during past decades, it has been applied to the TSC problems.
A three layers fully connected Convolutional Neural Network with an average pooling layer designed for time-series classification has been introduced in \cite{y10}. Fawaz et al. \cite{y3} proposed new data augmentation techniques in order to avoid overfitting. In \cite{y27}, the authors modified the cost function to enable the FCN model to be more sensitive with skewed time series data sets. In forecasting time series of spatial process, a dynamical spatio-temporal model was proposed in \cite{y28}. For limited time series data sets, like Electronic Health Records (EHRs) sequential data, Che et al. \cite{y29} recently trained a generative adversarial network with CNN to generate satisfactory risk prediction. 
Despite all these applications of deep learning in time series data sets across various domain fields, obstacles to further apply DNNs to other data resources still exist. After all, the fundamental challenge is still the availability of big data across different domains \cite{y30}.

On the other hand, transfer learning becomes heated research topic these days, and has also been applied to time series data mining tasks. In this case, a model is simultaneously learned both on the source and the target domains to minimize effects of cross-domain discrepancy within the learned model \cite{y31,y32}

In \cite{y33}, in time series anomaly detection, the authors proposed a transfer learning approach with a 1-NN DTW classifier to select time series transferred from the source to the target data set. 

Apart from time series classification tasks, transfer learning method was found in time series forecasting\cite{y34}, where the authors utilized a model trained on the historical wind-speed data of an old farm to predict wind speed in another farm. Also, similar techniques appeared in time series recognition tasks: a model was trained under similar conditions for acoustic phoneme recognition before it was applied to post-traumatic stress disorder diagnosis \cite{y35}.

Meanwhile, time series itself has been redefined, as sensor development enables improvements in model performance by taking into consideration multi-view time series data sets, which have received wide attention across many application domains during recent years. Because of their nice property that each of those multiple views provides support to each other and thus could enrich information on the object, it has been applied to many machine learning problems, such as clustering \cite{y36, y37}, classification \cite{y38, y39} on deep multi-view representation learning. 
	
To address these issues, several traditional classifiers developed for other fields have been modified to m.t.s. classification, such as SVM \cite{y40}, recurrent probabilistic neural network \cite{y41}, convolutional nonlinear component analysis\cite{y42}.

\section{Problem Definition}
In this section, we presented formal definitions for time series classification problem and multi-view time series classification problem. Let's start with the simplest definition of univariate time series.

\begin{Definition}
A univariate time series $X=\left[x_{1}, x_{2}, \ldots, x_{T}\right]$ is an ordered set of real values. The
length of $X$ is equal to the number of real values $T .$
\end{Definition}

We then define the multivariate time series.

\begin{Definition}
A multivariate time series $X=\left[x_{1}, \cdots, x_{m}\right] \in \mathbb{R}^{d \times m}$ is an ordered sequence of $d$ -dimensional vectors, in which $x_{i}$is the observation at the $i$-th timestamp, and $m$ is the length of time series.
\end{Definition}

Since the increasing number of various sensors, information of the same object could be collected from multiple perspectives. Here, we give a formal definition of multi-view multivariate time series.

\begin{Definition}
Multi-View Multivariate Time Series $(\mathrm{m} . \mathrm{t} . \mathrm{s} .) .$ A multi-view m.t.s $ \mathcal{X}=\left\{X_{(v)}\right\}, \quad v=1, \cdots, V$ is a set of time series data collected from multiple views, where $X_{(v)} \in \mathbb{R}^{d_{v} \times m_{v}}$ denotes the time series observed in the $v$ -th view, $d_{v}$ is the number of measurements of time series, $m_{v}$ is the length of time series, and $V$ is the total number of views.
\end{Definition}

Next we introduce the formal definition of Transfer Learning Based Multi-View Multivariate Time Series Classification.

\begin{Definition}
Transfer Learning Based Multi-View Multivariate Time Series Classification. Let $\tilde{\mathcal{X}}=\left\{X_{i, v} | i=1, \cdots, N, v=\right.$
$1, \cdots, V\}$ denote a set of multi-view $m$.t.s., where $N$ is the number of $m . t . s$ in each view, and $\mathcal{C}=\left\{C_{1}, \cdots, C_{c}\right\}$ denote
a set of class labels shared by $V$ views. Specifically, we define the last view $V$ is the target view and views from $1,\cdots,N-1$ all can be considered as source views. The task of transfer learning based multi-view m.t.s. classification is first to learn a classier from source views $\hat{\mathcal{X}} \backslash\left\{X_{i, V} | i=1, \cdots, N\right\},$ and then transfer the learned knowledge and features (the netwrok's weights) to the target view $\left\{X_{i, V} | i=1, \cdots, N\right\}$ and task.

For any set $\mathcal{X}=\left\{X^{(1)}, \ldots, X^{(J)}\right\},$ we denote set $\mathcal{X}$ with-
out the $j$ -th element as $\mathcal{X} \backslash\left\{X^{(j)}\right\}$
\end{Definition}

After reviewing all definitions above, it is natural to come up with the intuition that the framework of multi-view multivariate TSC may consist pieces of solutions to both multivariate TSC and univariate TSC tasks. Here we introduce another that we would applied in the following section. 

The technique of Importance Sampling (IS) can be
used to improve the sampling efficiency. The basic idea of the importance sampling is quite straightforward: instead of sampling from the nominal distribution, we draw samples from an alternative distribution from which an appropriate weight is assigned to each sample. Most recent works focus on its application on stochastic gradient descent. Now we apply this perspective to different views, we construct a dynamic inter-view importance metric to measure each source view's importance contributing to target view.

\begin{Definition}
Dynamic Inter-View Importance Score $\mathcal{I}$ is a kind of metric that indicates view $\mathcal{A}$'s importance to view $\mathcal{B}$ via sampled from the alternative distribution $Q$ (for this task this is the approximated posterior distribution of density in latent space) 
\end{Definition}

The notion of dynamic inter-view importance is similar to view-level similarity, to some degree, but with more uncertainty.

\section{Adaptive Transfer Learning of Multi-View Time Series Classification}

In this section we presented our novel approach. We first gave a brief introduction of our framework and then elaborated it in details later.

Since inter-view importance is hard to define on multi-view time series datasets, we proposed this dynamic measurement, which is constructed by the following steps. First we compute similarities between corresponding multivariate time series in different views. After carrying out the all the pairwise similarities, we put those similarities into a latent space and apply density estimation methods such as kernel density estimation or normalizing flow to approximate the posterior similarity distributions. Then, we look at it from a perspective of importance sampling, where this sampled value from inter-view importance approximated posterior distribution could be viewed as importance value for the target view, which would control the knowledge transfer degree of each source view in the pre-training process. 

\subsection{Computation of Inter-View Importance Value}

To implement transfer learning on multi-view time series data, we need to capture inter-view importance. In this part, we list various measurements. Among them, we chose Dynamic Time Wrapping (DTW) and Bag of SFA Symbols (BOSS) to calibrate the inter-view importance, as these two measure show great performance in univariate time series cases. We speculate the performance to be consistent. 

Complete procedures to carry out inter-view importance value are as follows.

Let $\mathcal{S}$ be the source view $\mathcal{S} = \left\{X_{i, S} | i=1, \cdots, N\right\}$ , and target view be  $\mathcal{T} = \left\{X_{i, T} | i=1, \cdots, N\right\}$. The order from $1$ to $N$ in both source and target view corresponds to each other. In other words, for the $j-th$ term, we have both views $X_{j, S}$ and $X_{j, T}$.

Notice that $X_{j, S}$ and $X_{j, T}$ are multivariate time series, therefore when calculating similarities we should decompose multivariate time series $X_{j, S}$ and $X_{j, T}$ into univariate time series $\mathcal{X}_{j, S} = \left\{X_{k, j, S} | k=1, \cdots, K\right\}$ and $\mathcal{X}_{j, T} = \left\{X_{k, j, T} | k=1, \cdots, K\right\}$ , where $K$ denotes the dimension of multivariate time series in both views.

We then compute the corresponding decomposed pairwise univariate time series distance between set $\mathcal{X}_{j, S}$ and $\mathcal{X}_{j, T}$ 
\begin{equation}
    S_{k,j} = dist(X_{k, j, S} ,X_{k, j, T} )
\end{equation}
Here, we list several widely-used time series distance measurements, but remember we chose Dynamic Time Wrapping (DTW) and Bag of SFA Symbols (BOSS) for in this paper.
\begin{itemize}

    \item \textbf{Dynamic Time Wrapping (DTW):} could yield better performance when the lengths differ. Dynamic time warping distance is given as
\begin{equation}
d_{DTW}({\mathbf{X}_{T},\mathbf{Y}_{T}})={min}_{r \in M}(\sum_{i=1,...,m} |X_{a_{i}}-Y_{b_{i}}|)
\end{equation}
where $M$ represents the set of all possible sequences of $m$ pairs. Under most circumstances, shaped-based approaches give better results on small-scale time series data sets with much less noises and outliers.
    

    \item \textbf{Bag of SFA Symbols (BOSS):} BOSS uses windows to form words over series. BOSS uses a truncated Discrete Fourier Transform (DFT) instead of PAA\cite{y22} on each window, and the truncated series is discretized through a technique called Multiple Coefficient Binning (MCB). Then it windows each series to form word distribution through the application of DFT and discretization by MCB.
\end{itemize}

Every decomposed univariate time series distance between source and target view is a good measure of importance value, and all of these measurements are later transformed to $K$ dimensions, adjusted by length of multivariate time series and the number of time series views. 
\begin{equation}
    S_{j}=\left(S_{1,j},\cdots,S_{K,j}\right)
\end{equation}
We carry out all above pairwise distance $S_{1}$ to $S_{N} $, and put all values into this observation set $\mathcal{S} = \left\{S_{1},S_{2},\cdots,S_{N}\right\}$ in latent space. Now, after mapping all elements into a latent space $\mathcal{\kappa}$, we are ready to construct a approximated posterior distribution of importance values.

\subsection{Density Estimation}

In this part, we approximated a posterior distribution $Q(S)$ to describe the importance relationships between source views and target view. Below, we separately elaborated on how to deal with high-dimensional and low-dimensional scenarios.
\begin{itemize}

\item In a high-dimensional scenario, a normalizing flow model $f$ is constructed as an invertible transformation which maps observed data points in latent space $\mathcal{S} = \left\{S_{1},S_{2},\cdots,S_{N}\right\}$ by a standard Gaussian latent variable $\mathbf{h}=\mathcal{F}(\mathbf{S}),$ as what is like in non-linear Independent Component Analysis. Stacking individual simple invertible transformations is the key idea in designing a flow model. Explicitly, $\mathcal{F}$ is made up of a series of invertible flow $\mathcal{F} = f_L \circ \cdots \circ f_1$, with each $f_{i}$ having a tractable Jacobian determinant. This way, sampling is efficient, as can be performed by computing $\mathcal{F}^{-1} = f^{-1}_1 \circ \cdots \circ f^{-1}_l$ for $\mathbf{S} \sim \mathcal{N}(\mathbf{0}, \mathbf{I})$. So is the training process by maximum likelihood. Because the model density is easy to compute and differentiate with respect to the flows $f_{i} .$ \cite{y43}

\begin{equation}
\log Q(\mathbf{S})=\log \mathcal{N}(\mathcal{F}(\mathbf{S}) ; \mathbf{0}, \mathbf{I})+\sum_{i=1}^{L} \log \left|\operatorname{det} \frac{\partial f_{i}}{\partial f_{i-1}}\right|
\end{equation} In computing the Jacobian determinant, we set a threshold for adding stochastic perturbation $\epsilon$ to balance the computational complexity and precision, in case the matrices are singular. The trained flow model can be considered as maximizing a posterior estimation $Q(\mathbf{S})$ for inter-view importance value in latent space.

\item In low-dimensional scenarios, kernel density estimation is a good fit. The univariate kernel density estimator for a continuous variable based on a sample $\left\{S_{1}, \ldots, S_{N}\right\}$ at the evaluation point $y$ can be expressed as

\begin{equation}
 Q(S)=\frac{1}{n} \sum_{i=1}^{n} w\left(S-S_{i} ; h\right)   
\end{equation}where $w(\cdot)$ is the kernel function, a symmetric weightion, and $h$ is the smoothing parameter or bandwidth. 

For multivariate kernel density estimation, let  $\left\{S_{1},S_{2},\cdots,S_{N}\right\}$ be a sample of d-variate random vectors drawn from a common distribution generated by density function ƒ. The kernel density estimate is
\begin{equation}
Q(\mathbf{S})=\frac{1}{n} \sum_{i=1}^{n} K_{\mathbf{H}}\left(\mathbf{S}-\mathbf{S}_{i}\right)
\end{equation}
where $\mathrm{H}$ is the bandwidth (or smoothing), $d \cdot d$ matrix which is symmetric and positive definite, and $K_{\mathbf{H}}$, the kernel function, can be further written with respect to $\mathbf{S}$
\begin{equation}
K_{\mathbf{H}}(\mathbf{S})=|\mathbf{H}|^{-1 / 2} K\left(\mathbf{H}^{-1 / 2} \mathbf{S}\right)
\end{equation}
which is plainly a symmetric multivariate density. For simplicity purposes, we directly choose the standard multivariate normal kernel,
\begin{equation}
K_{\mathbf{H}}(\mathbf{S})=(2 \pi)^{-d / 2}|\mathbf{H}|^{-1 / 2} e^{-\frac{1}{2} \mathbf{S}^{\mathrm{T}} \mathbf{H}^{-1} \mathbf{S}}
\end{equation}

\end{itemize}

\begin{algorithm}[!ht]
\caption{Adaptive Transfer Learning of Multi-View Time Series Classification}
\begin{algorithmic}[1]
\FOR{Each corresponding multivariate time series pair $X_{j,S}$ and $X_{j,T}$ in source view $\mathbf{X}_{S}$and target view $\mathbf{X}_{T}$}
\STATE Decompose $X_{j,S}$ and $X_{j,T}$ into 
\begin{equation*}
    \mathcal{X}_{j, S} = \left\{X_{k, j, S} | k=1, \cdots, K\right\}
\end{equation*}
\begin{equation*}
    \mathcal{X}_{j, T} = \left\{X_{k, j, T} | k=1, \cdots, K\right\}
\end{equation*}
\STATE Compute univariate time series similarity by
\begin{equation*}
    S_{k,j} = dist(X_{k, j, S} ,X_{k, j, T} )
\end{equation*}
\STATE Collect univariate similarity and update to latent space
\begin{equation*}
    S_{j}=\left(S_{1,j},\cdots,S_{K,j}\right)
\end{equation*}
\begin{equation*}
    \mathcal{S} = \left\{S_{1},S_{2},\cdots,S_{N}\right\}
\end{equation*}
\ENDFOR

\IF{$\mathcal{S}$ is High-dimensional}
\STATE Construct posterior distribution by maximizing $\log Q(\mathbf{S})$ on $\{{S}_1,\cdots,{S}_N\}$
\ENDIF

\IF{$\mathcal{S}$ is Low-dimensional}
\STATE Construct posterior distribution by kernel density estimation
\begin{equation*}
Q(\mathbf{S})=\frac{1}{n} \sum_{i=1}^{n} K_{\mathbf{H}}\left(\mathbf{S}-\mathbf{S}_{i}\right)
\end{equation*}
\ENDIF
\STATE Draw $m$ samples for ${N}(\mathbf{0}, \mathbf{I})$ then go through $Q(\mathbf{S})$
\begin{equation*}
    S_{i} \sim \mathcal{N}(\mathbf{0}, \mathbf{I})
\end{equation*}
\begin{equation*}
    S^{t}_{i} = Q(S_{i})
\end{equation*}
\STATE Compose matrix $S^{t}_{I.V.}$
\STATE Calculate matrix norm
\begin{equation*}
\|g_{t}\| = \| S^{t}_{I.V.} \|
\end{equation*}
\end{algorithmic}
\end{algorithm}

\subsection{Importance Value Sampling and Matrix Norm Computation}

After constructing a posterior distribution, we could conduct importance value sampling now. We start by sampling importance values in mini-batches. Let $\mathbf{S}=\{S^{t}_{1},S^{t}_{2},...,S^{t}_{m}\}$ denotes the batch with size $m$, $\mathcal{N}(\mathbf{0}, \mathbf{I})$ denotes the standard Gaussian distribution, and we have
\begin{equation}
    S \sim \mathcal{N}(\mathbf{0}, \mathbf{I})
\end{equation}
Then, let $\mathbf{S}$ go through the trained normalizing flow $Q(\mathbf{S})$. If low-dimensional scenario, we draw $m$ samples from approximated kernel distribution. In this way, we can acquire dynamic inter-view importance values in new batches on approximated distribution. Put all the sampled vectors into a matrix, we have
\begin{equation}
    S^{t}_{I.V.} = \left[\begin{array}{c}{S^{t}_{1}}\\ {\vdots} \\ {S^{t}_{m}} \end{array}\right]
\end{equation}
If we unfold each $S^{t}_{i}$, we have 

\begin{equation}
    S^{t}_{I.V.} = \left[\begin{array}{ccc}{S^{t}_{1,1}} & {\cdots} & {S^{t}_{K,1}}\\ {\vdots} & {\ddots} & {\vdots} \\ {S^{t}_{1,m}} & {\cdots} & {S^{t}_{K,m}} \end{array}\right]
\end{equation}
Here we compute the norm per matrix. The elements of these sample-composed matrices contain the information of importance value in every dimension, which can be accumulated and computed via matrix norm.

Then we compute the matrix norm of $S^{t}_{I.V.}$ by
\begin{equation}
g_{t}=\left\|S_{I.V.}^{t}\right\|
\end{equation}
Finally, we arrive at the output $g_{t}$ as the desired probablistic representation of the inter-view importance values between source view  $\mathcal{S} = \left\{X_{i, S} | i=1, \cdots, N\right\}$ and target view $\mathcal{T} = \left\{X_{i, T} | i=1, \cdots, N\right\}$, which describes the degree of knowledge transfer in the pre-training process. We will elaborate how we use these dynamic importance values to control the the degree of knowledge transfer in the experiment section.

\subsection{Model Architecture}

We selected one dimensional Fully Convolutional Neural Network\cite{y10} (FCN) and  Long Short-Term Memory (LSTM)\cite{y3} network model to construct our adaptive transfer learning framework. The reason behind our choice is these networks' robustness as they have already achieved state-of-the-art results on several data sets from the UCR archive and UEA repository. However, please note that our adaptive transfer learning framework is totally independent of the chosen neural networks.


\begin{table}[!t]
\newcommand{\tabincell}[2]{\begin{tabular}{@{}#1@{}}#2\end{tabular}}
\centering
\begin{tabular}{c}
\hline
\hline
LSTM  ( 256)\\ 
\hline
DenseLayer (classes) \\
\hline
softmax\\
\hline
\hline
\end{tabular}
\caption {The structure of Long Short-Term Memory Recurrent Neural Network (LSTM-RNN)}
\label{net_structure}
\end{table}

\begin{table}[!t]
\newcommand{\tabincell}[2]{\begin{tabular}{@{}#1@{}}#2\end{tabular}}
\centering
\begin{tabular}{c}
\hline
\hline
Conv1D (length = $1$ or $8$) $\times$ 128\\ 
\hline
BN+relu+Dropout(0.2)\\
\hline
Conv1D (length = $1$ or $5$) $\times$ 256\\
\hline
BN+relu+Dropout(0.2)\\
\hline
Conv1D (length = $1$ or $3$) $\times$ 128\\
\hline
BN+relu+Dropout(0.2)\\
\hline
DenseLayer(classes)\\
\hline
softmax\\
\hline
\hline
\end{tabular}
\caption {The structure of Fully Convolutional Network (FCN). (In case of overfitting, we set the kernel size =1 when varying on Movement-AAL data set for reducing of parameters.)}
\label{net_structure}
\end{table}

\begin{table}[!t]
\newcommand{\tabincell}[2]{\begin{tabular}{@{}#1@{}}#2\end{tabular}}
\centering
\begin{tabular}{c}
\hline
\hline
DenseLayer(128) \\ 
\hline
relu\\
\hline
DenseLayer (128) \\ 
\hline
relu\\
\hline
DenseLayer (classes) \\ 
\hline
softmax\\
\hline
\hline
\end{tabular}
\caption {The structure of Multilayer Perceptron (MLP).}
\label{net_structure}
\end{table}

\begin{table*}[h]
\centering
\begin{tabular}{l|c|c|c}
\hline
\hline
    \multicolumn{1}{c|}{} &
    \multicolumn{1}{c|}{Daily and Sports Activity}         & \multicolumn{1}{c|}{ Movement } & \multicolumn{1}{c}{Self-Regulation of SCPs} 
    \\ \hline
FCN     &   0.76842   & 0.61538 & 0.70307   \\
DTW-FCN    & \textbf{0.79342}       & 0.68269   & 0.78840  \\
BOSS-FCN  & 0.78815            & \textbf{0.684262}   & \textbf{0.79522}   \\
\hline
LSTM  &       0.41711           & 0.53846   & 0.45051 \\
DTW-LSTM  &     \textbf{0.48378}     & 0.58653 & \textbf{0.60409}   \\
BOSS-LSTM &  0.43509           & \textbf{0.58854}   & 0.59727  \\
\hline
MLP  &    0.94123     &0.55769 & 0.70307\\
DTW-MLP  &    \textbf{0.94737}  &  \textbf{0.58654}  & \textbf{0.76451}\\
BOSS-MLP    & 0.94693 &0.58121 &0.75768\\

\hline

\end{tabular}

\caption {Validating classification accuracy of baseline and proposed approaches with Dynamic Time Warping (DTW) and Bags of SFA Symbols (BOSS) measurement on 'Daily and Sports Activity', 'Movement' and 'Self-Regulation of SCPs'}
\label{table:1a}
\end{table*}

\section{Experiment Result}

\subsection{Data Set}
\begin{itemize}
    \item \textbf{UCI Daily and Sports Activity Data Set} \cite{y44} contains motion sensor data of 19 daily and sports activities, each of which is performed by 8 subjects within 5 minutes. In particular, the subjects were asked to perform these activities in there own styles without any restrictions. As a result, the time series samples for each activity have considerable inter-subject variations in terms of speed and amplitude, which makes it extremely difficult to reach accurate classification results. During the data collection, nine sensors were put onto torso, right arm, left arm, right leg, and left leg these five units. The 5-minute time series collected from each subject is divided into 5-second segments. For each activity, the total number of segments is 480, and each segment is considered as a multivariate time series sample of size 45 × 125.

    \item \textbf{Indoor User Movement Prediction from RSS Data Set} represents a real-life benchmark in the area of Ambient Assisted Living applications. The binary classification task consists in predicting the pattern of user movements in real-world office environments from time-series generated by a Wireless Sensor Network (WSN). Input data contains temporal streams of radio signal strength (RSS) measured between the nodes of a WSN, comprising 5 sensors: 4 anchors deployed in the environment and 1 mote worn by the user. Data has been collected. In the provided data set, the RSS signals have been re-scaled to the interval [-1,1], singly on the set of traces collected from each anchor. Target data consists in a class label indicating whether the user's trajectory will lead to a change in the spatial context (i.e. a room change) or not. \cite{}
    
    \item \textbf{Self-Regulation of SCPs Data Set} was taken from a healthy subject. The subject was asked to move a cursor up and down on a computer screen. During the recording, the subject received visual feedback of his slow cortical potentials.  Cortical positivity lead to a downward movement of the cursor on the screen.  Cortical negativity lead to an upward movement of the cursor.  Each trial lasted 6s. During every trial, the task was visually presented by a highlighted goal at either the top or bottom of the screen to indicate negativity or positivity from second 0.5 until the end of the trial. The visual feedback was presented from second 2 to second 5.5. Only this 3.5 second interval of every trial is provided for training and testing and 896 samples per channel for every trial.
\end{itemize}

\subsection{Experiment Setup}

\begin{itemize}
    \item For UCI Daily and Sports Activity data sets, we regard information from sensors on different parts the body as different views and thus this data set gives 5 views. Then, we randomly pick 4 out of the 5 as source views and the rest as target view. Next, we set 6 out of 8 subjects as training set and the other 2 subjects as testing set. Based on the fact that multivariate time series in different views are all 9-dimensional, we select the high-dimensional solution (Normalizing Flow) for latent space density estimation after computing inter-view importance. After acquiring the $g-{t}$ for 4 source views, we consider these 4 value as the importance score from corresponding source view to target view and control the proportion of pre-training. We set 200 epochs for all these 4 source views, and each takes up a proportion $g_{ti}$ of their corresponding importance score. For loss function, we choose categorical entropy.
    \begin{equation}
        g_{ti} = \frac{\|g_{ti}\|}{\sum_{j=1}^{4} \|g_{tj}\|}\cdot T
    \end{equation}
    where $T$ denotes total epoch for pretraining and $g_{ti}$ denotes the assigned number of epoch for a specific source view $\mathcal{i}$.
    \item Indoor User Movement Prediction from RSS Data Set contains 4 sensors. By considering them as 4 different views, we get 4 views, but with different timestamp (in this case average is 42). 
    \begin{itemize}
        \item Pad the shorter sequences with zeros to make the length of all the series equal.
        \item Find maximum time series length and pad the rest of the shorter sequences with last-row values.
        \item Identify minimum time series length of each data set and truncate all the other series to that length. However, this leads to a huge information loss.
        \item Calculate average series lengths, truncate all longer-than-average series, and pad all shorter-than-average series.
    \end{itemize}
    After these pre-processing procedures, we randomly select 3 out of 4 views as source views and the rest as target view. By applying model described above, we can calculate $g_{t}$ to get a desired probablistic representation of corresponding importance scores. We set 120 epochs for pre-training. The detailed algorithm is the same with UCI Daily and Sports Activity data sets above.
    
    \item For Self-Regulation of SCPs data set, we take time series from 6 channels as a 6-view time series data set. We randomly pick 5 out of 6 views as source views and the rest as the target view, and then apply proposed model to get $g_{t}$. With this probablistic representation, we finally get each one's corresponding importance score. We set 100 epochs for pre-training. The detailed algorithm for assigning weight during pretraining is the same with UCI Daily and Sports Activity data sets above.
    
\end{itemize}

For all the network training, we normally applied batch size as $64$ and optimizer as AdaM. Due to the error and randomness, we repeat every experiment with 5 times and compute average classification accuracy.

\subsection{Result Analysis}

We run 3 baseline models (Long Short-Term Memory Recurrent Neural Network (LSTM-RNN), Fully Convolutional Network (FCN) and Multi-Layer Perceptron (MLP)) and 6 adaptive transfer learning frameworks (Dynamic Time Wraping (DTW)-LSTM, Bag Of SFA Symbols (BOSS)-LSTM, DTW-FCN, BOSS-FCN, DTW-MLp and BOSS-MLP) on these 3 data sets. Our fine-tuned classification accuracy results are shown in the table and related figures. 

As shown in Figure.2,3 and 4, in most scenarios, our proposed approaches perform better results than baselines. Due to the pretrain process, our proposed approaches always reach a high classification accuracy at beginning, but the proposed approaches take the lead in classification accuracy all the time during network training. As listed results in table.4, proposed approaches reach a better accuracy after the training process. We provide the density estimation results of latent space of different datasets too.

\section{Conclusion}

In this paper, we presented an adaptive transfer learning framework on multi-view multivariate time series data. We looked at the multi-view time series data through a perspective of importance sampling, where we attempted to measure the importance value of a specific source view time sequence for the target view time sequence. The inter-view importance was carried out in the following procedures. First, we calculated decomposed corresponding pairwise univariate time series distance. Second, we updated the importance value into a latent space to calculate observation density estimation. Finally, we arrived at approximated posterior distribution. Particularly, we discussed two scenarios when input dimensions are either high or low. Later, we sampled several importance values to compute a composed matrix norm as output importance score, which also indicates the degree of knowledge transfer in the pre-training process. On average, our proposed adaptive transfer learning framework demonstrates a generally improved classification performance of $2\%$ to $5\%$ over some state-of-the-art baseline models.

\begin{figure*}[ht]
    \centering
    \subfigure[Accuracy on SCPs validation set, w and w/o boss.]
    {
    \begin{minipage}{5cm}
    \centering
    \includegraphics[width=5cm]{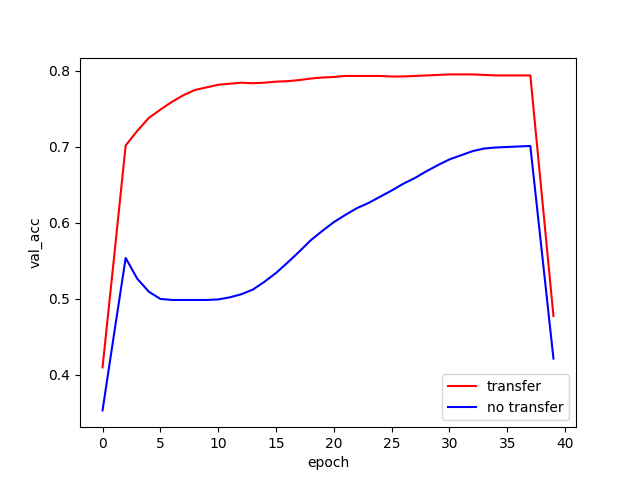}
    \label{focal_loss}
    \end{minipage}%
    }
    \subfigure[Accuracy on Movement validation set, w and w/o boss.]
    {
    \begin{minipage}{5cm}
    \centering
    \includegraphics[width=5cm]{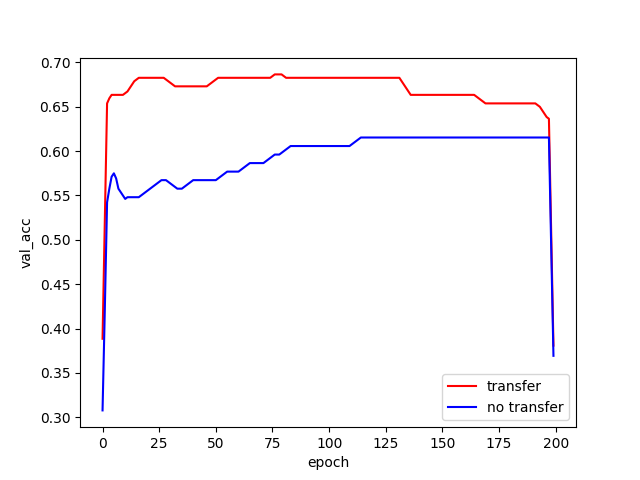}
    \label{focal_loss+MSE_loss}
    \end{minipage}%
    }
    \subfigure[Accuracy on Sports validation set, w and w/o boss.]
    {
    \begin{minipage}{5cm}
    \centering
    \includegraphics[width=5cm]{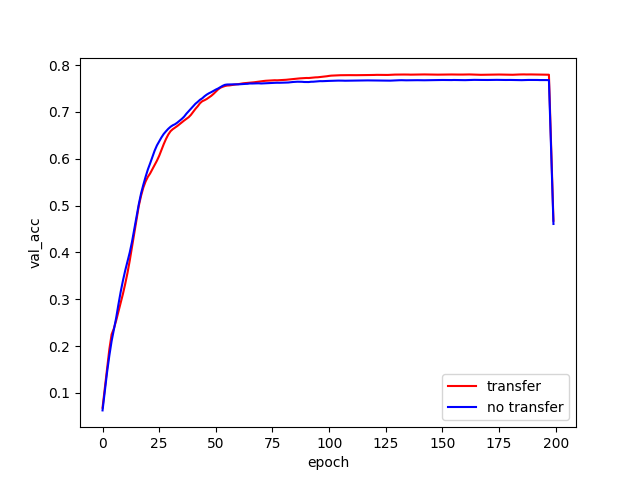}
    \label{focal_loss}
    \end{minipage}%
    }
     \centering
    \subfigure[Accuracy on SCPs validation set, w and w/o dtw.]
    {
    \begin{minipage}{5cm}
    \centering
    \includegraphics[width=5cm]{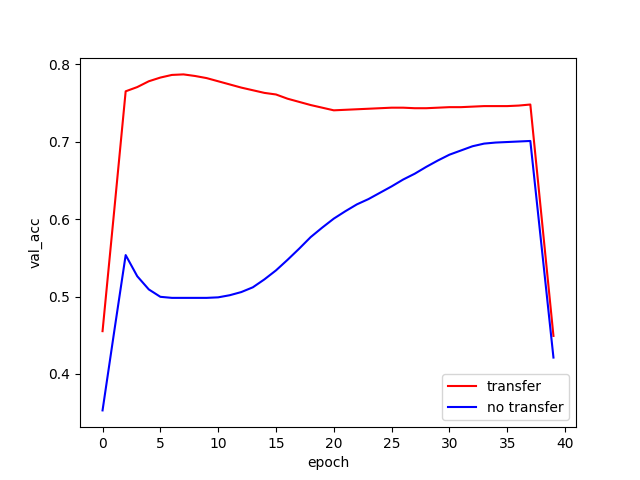}
    \label{focal_loss}
    \end{minipage}%
    }
    \subfigure[Accuracy on Movement validation set, w and w/o dtw.]
    {
    \begin{minipage}{5cm}
    \centering
    \includegraphics[width=5cm]{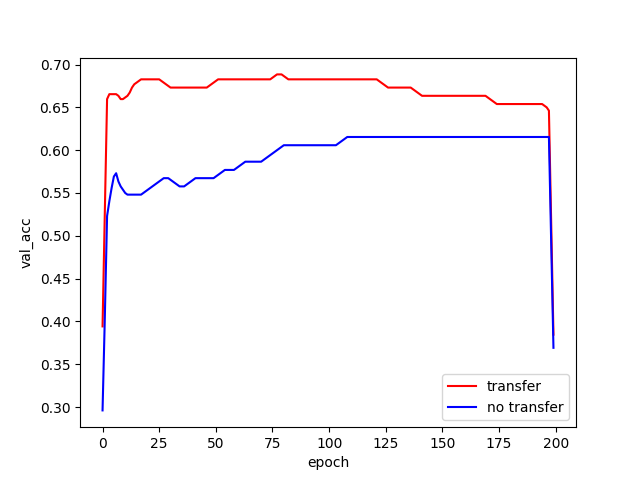}
    \label{focal_loss+MSE_loss}
    \end{minipage}%
    }
    \subfigure[Accuracy on Sports validation set, w and w/o dtw.]
    {
    \begin{minipage}{5cm}
    \centering
    \includegraphics[width=5cm]{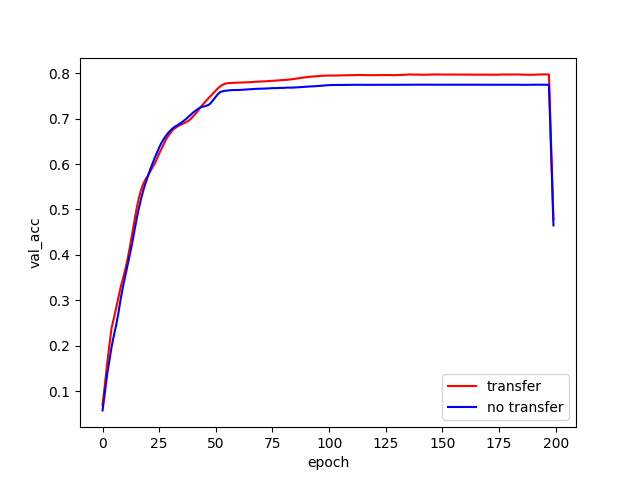}
    \label{focal_loss}
    \end{minipage}%
    }
\caption{Final results on Fully Convolutional Network (FCN).}
\label{Figure:Loss-comparsion}
\end{figure*}

\begin{figure*}[h]
    \centering
    \subfigure[Accuracy on SCPs validation set, w and w/o boss.]
    {
    \begin{minipage}{5cm}
    \centering
    \includegraphics[width=5cm]{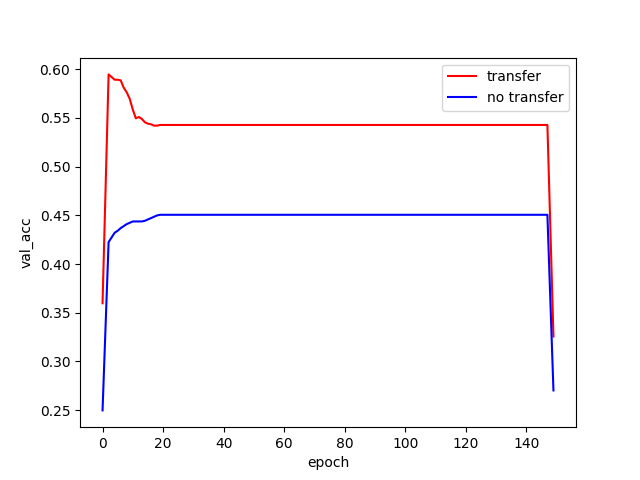}
    \label{focal_loss}
    \end{minipage}%
    }
    \subfigure[Accuracy on Movement validation set, w and w/o boss.]
    {
    \begin{minipage}{5cm}
    \centering
    \includegraphics[width=5cm]{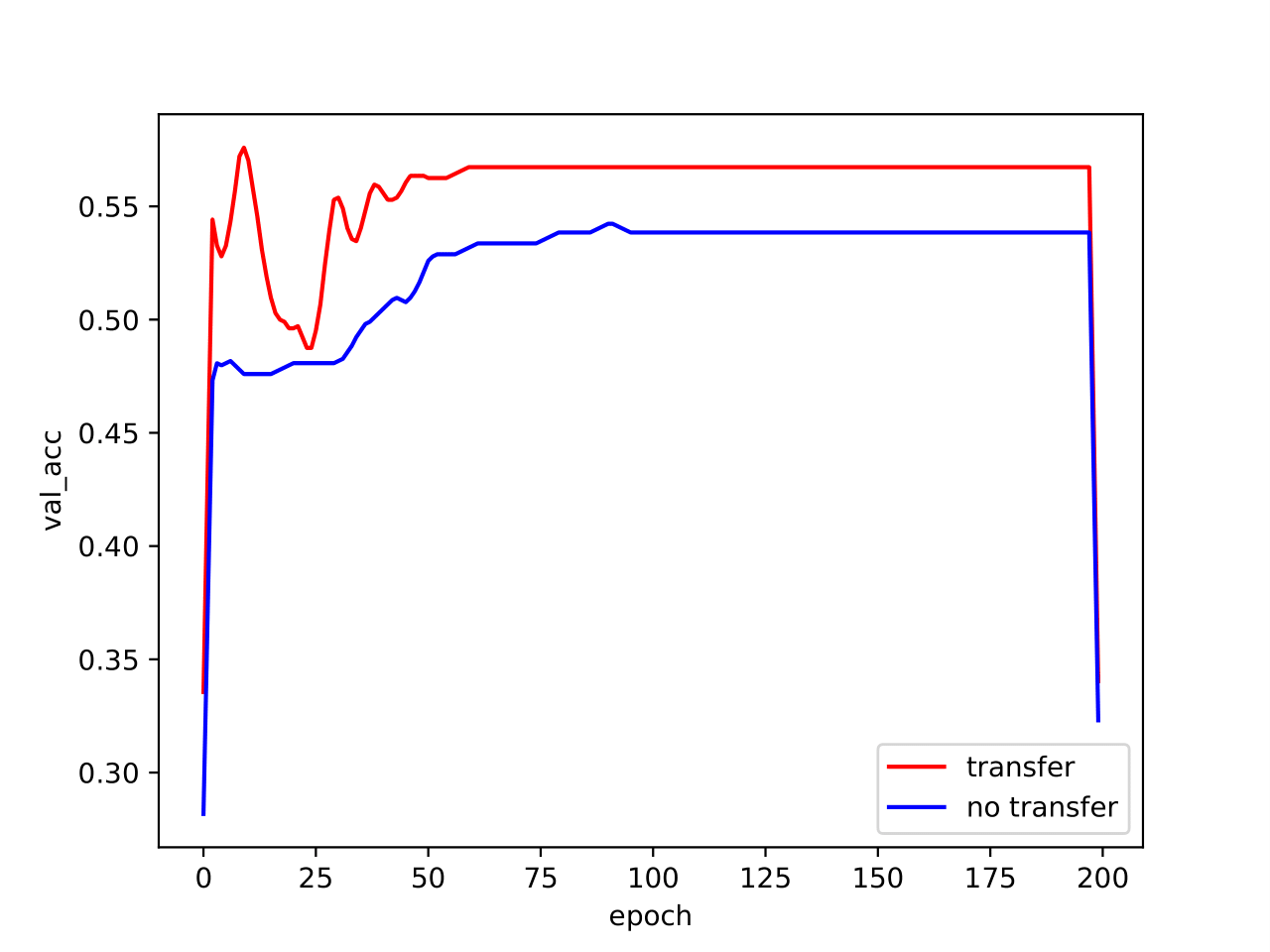}
    \label{focal_loss+MSE_loss}
    \end{minipage}%
    }
    \subfigure[Accuracy on Sports validation set, w and w/o boss.]
    {
    \begin{minipage}{5cm}
    \centering
    \includegraphics[width=5cm]{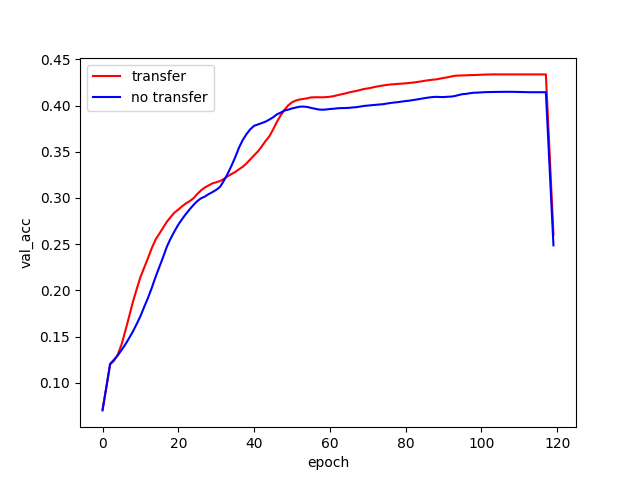}
    \label{focal_loss}
    \end{minipage}%
    }
     \centering
    \subfigure[Accuracy on SCPs validation set, w and w/o dtw.]
    {
    \begin{minipage}{5cm}
    \centering
    \includegraphics[width=5cm]{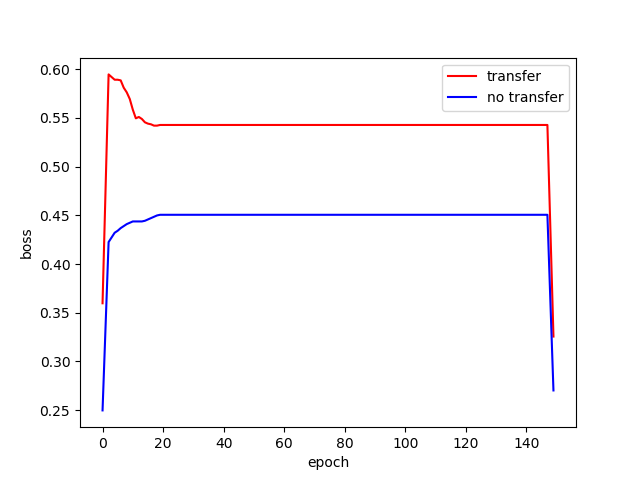}
    \label{focal_loss}
    \end{minipage}%
    }
    \subfigure[Accuracy on Movement validation set, w and w/o dtw.]
    {
    \begin{minipage}{5cm}
    \centering
    \includegraphics[width=5cm]{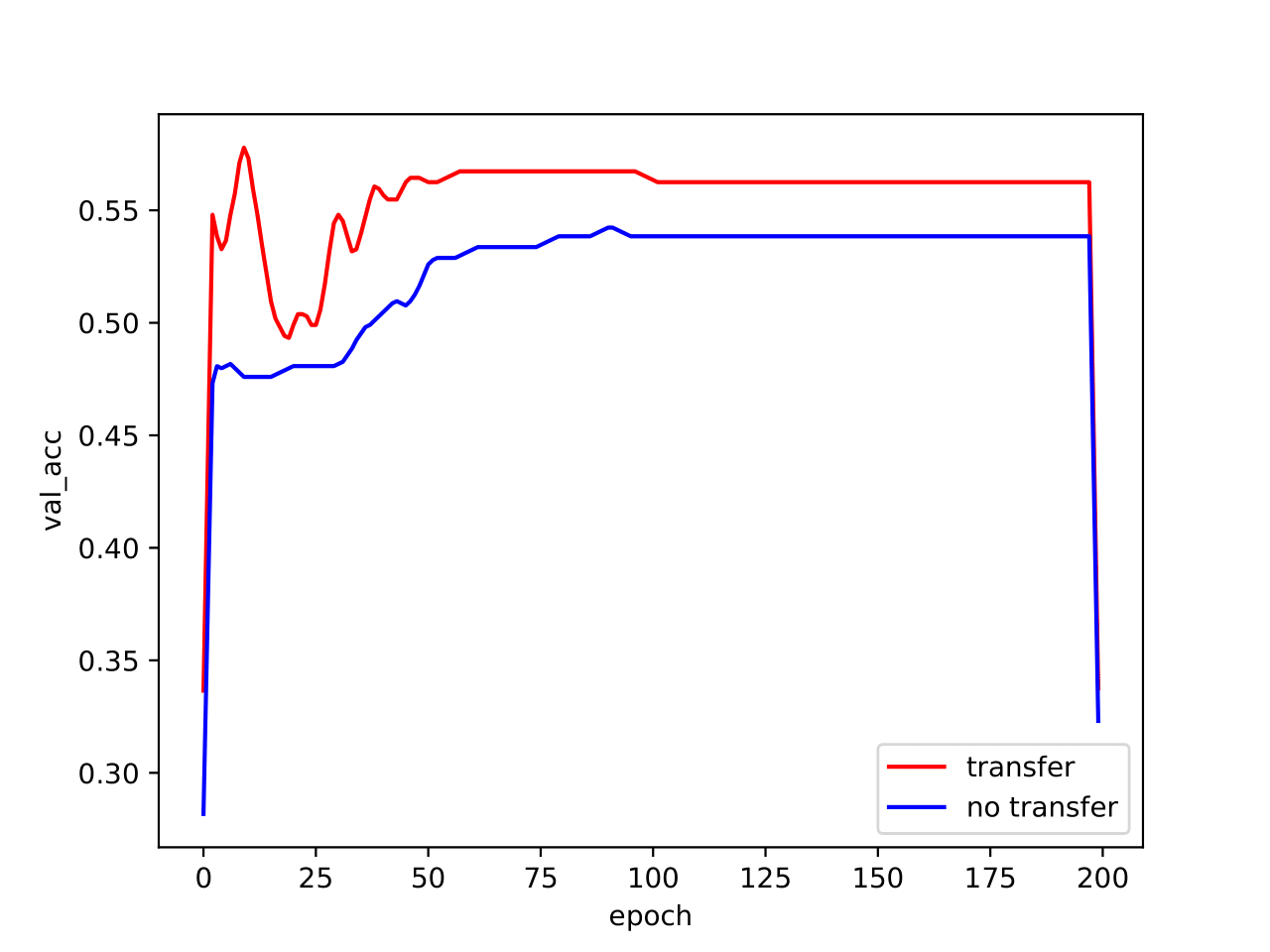}
    \label{focal_loss+MSE_loss}
    \end{minipage}%
    }
    \subfigure[Accuracy on Sports validation set, w and w/o dtw.]
    {
    \begin{minipage}{5cm}
    \centering
    \includegraphics[width=5cm]{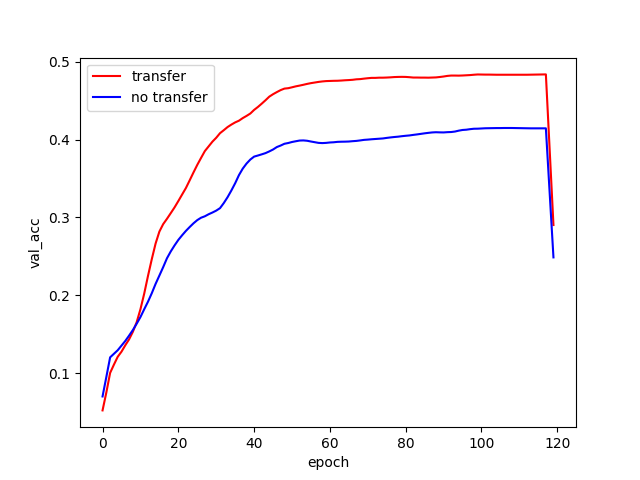}
    \label{focal_loss}
    \end{minipage}%
    }
    
\caption{Final results on Long Short-Term  Memory Recurrent Neural Network (LSTM-RNN).}
\label{Figure:Loss-comparsion}
\end{figure*}

\begin{figure*}[h]
    \centering
    \subfigure[Accuracy on SCPs validation set, w and w/o boss.]
    {
    \begin{minipage}{5cm}
    \centering
    \includegraphics[width=5cm]{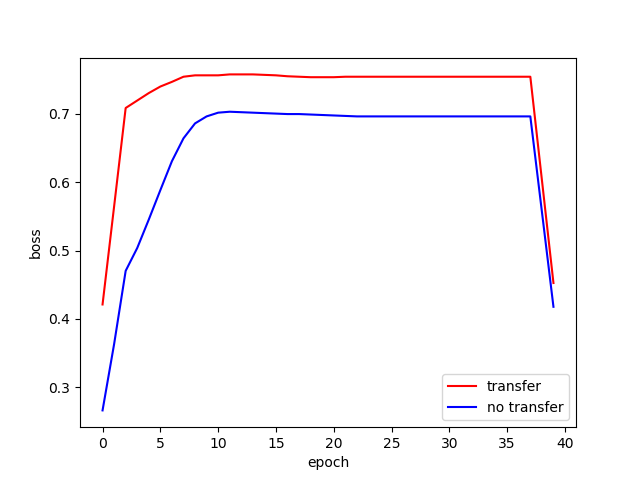}
    \label{focal_loss}
    \end{minipage}%
    }
    \subfigure[Accuracy on Movement validation set, w and w/o boss.]
    {
    \begin{minipage}{5cm}
    \centering
    \includegraphics[width=5cm]{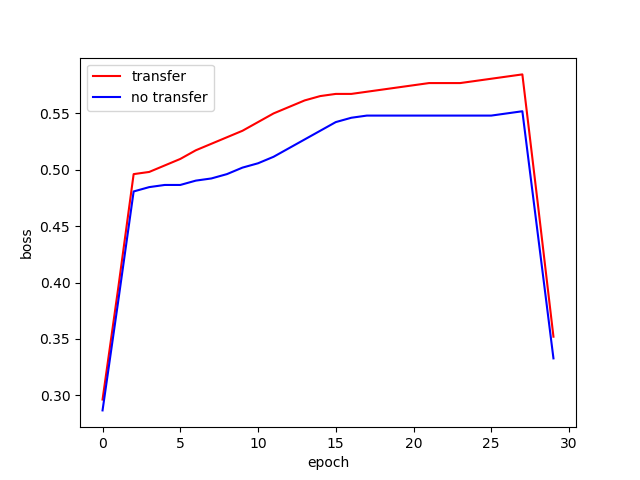}
    \label{focal_loss+MSE_loss}
    \end{minipage}%
    }
    \subfigure[Accuracy on Sports validation set, w and w/o boss.]
    {
    \begin{minipage}{5cm}
    \centering
    \includegraphics[width=5cm]{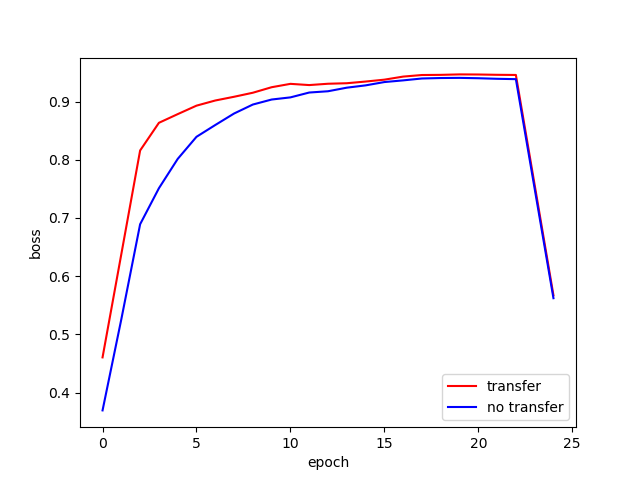}
    \label{focal_loss}
    \end{minipage}%
    }
     \centering
    \subfigure[Accuracy on SCPs validation set, w and w/o dtw.]
    {
    \begin{minipage}{5cm}
    \centering
    \includegraphics[width=5cm]{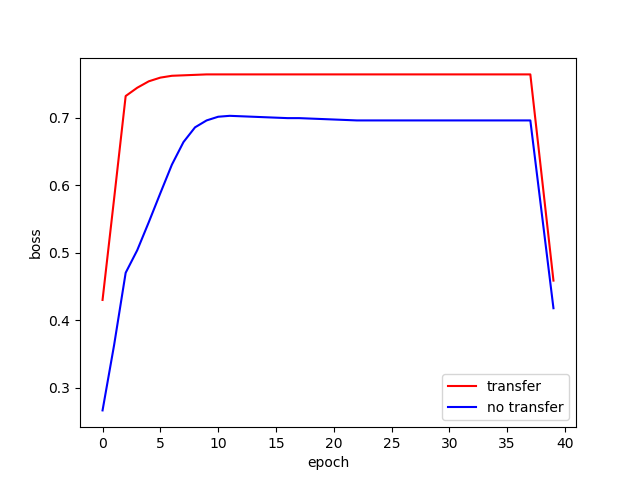}
    \label{focal_loss}
    \end{minipage}%
    }
    \subfigure[Accuracy on Movement validation set, w and w/o dtw.]
    {
    \begin{minipage}{5cm}
    \centering
    \includegraphics[width=5cm]{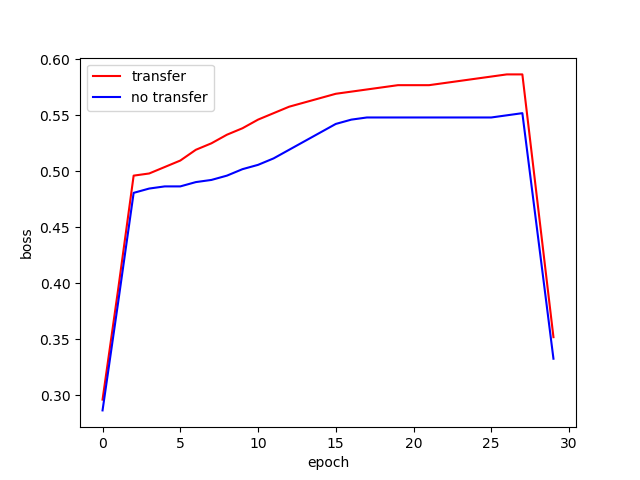}
    \label{focal_loss+MSE_loss}
    \end{minipage}%
    }
    \subfigure[Accuracy on Sports validation set, w and w/o dtw.]
    {
    \begin{minipage}{5cm}
    \centering
    \includegraphics[width=5cm]{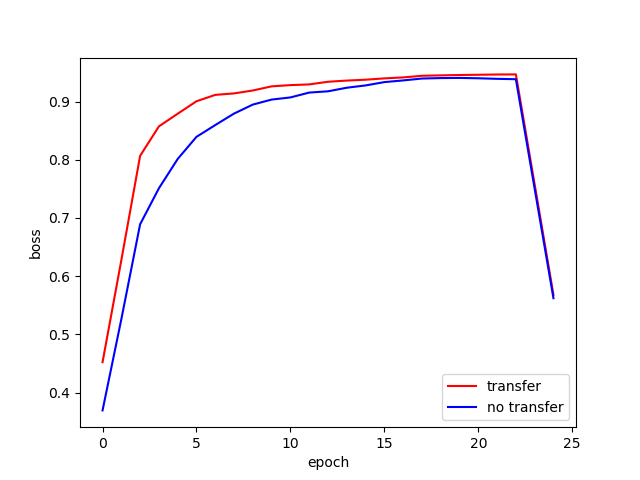}
    \label{focal_loss}
    \end{minipage}%
    }
\caption{Final results on  Multilayer Perceptron (MLP).}
\label{Figure:Loss-comparsion}
\end{figure*}

\begin{figure*}[ht]
    \centering
    \subfigure[Density Estimation of Inter-View Importance Latent Space via BOSS of Self-Regulation of SCPs Datasets(Between source view 1 and target view )]
    {
    \begin{minipage}{5cm}
    \centering
    \includegraphics[width=5cm]{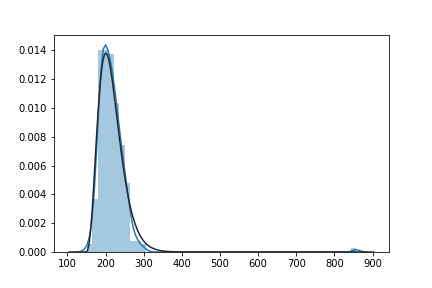}
    \label{focal_loss}
    \end{minipage}%
    }
    \subfigure[Density Estimation of Inter-View Importance Latent Space via BOSS of Self-Regulation of SCPs Datasets(Between source view 2 and target view )]
    {
    \begin{minipage}{5cm}
    \centering
    \includegraphics[width=5cm]{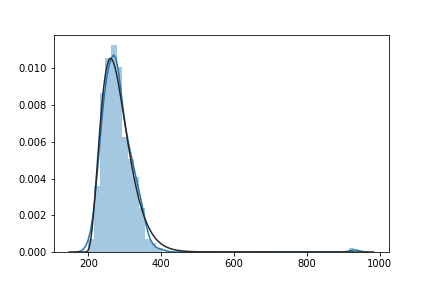}
    \label{focal_loss+MSE_loss}
    \end{minipage}%
    }
    \subfigure[Density Estimation of Inter-View Importance Latent Space via BOSS of Self-Regulation of SCPs Datasets(Between source view 3 and target view )]
    {
    \begin{minipage}{5cm}
    \centering
    \includegraphics[width=5cm]{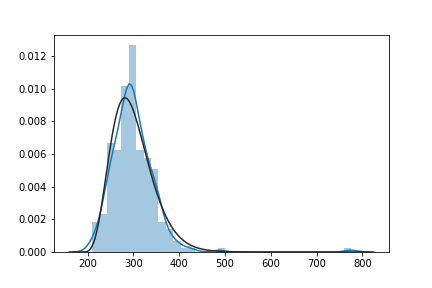}
    \label{focal_loss}
    \end{minipage}%
    }

     \centering
    \subfigure[Density Estimation of Inter-View Importance Latent Space via BOSS of Self-Regulation of SCPs Datasets(Between source view 4 and target view )]
    {
    \begin{minipage}{5cm}
    \centering
    \includegraphics[width=5cm]{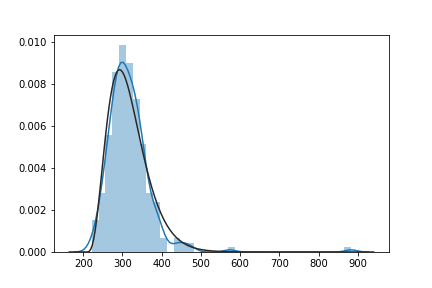}
    \label{focal_loss}
    \end{minipage}%
    }
    \subfigure[Density Estimation of Inter-View Importance Latent Space via BOSS of Self-Regulation of SCPs Datasets(Between source view 5 and target view )]
    {
    \begin{minipage}{5cm}
    \centering
    \includegraphics[width=5cm]{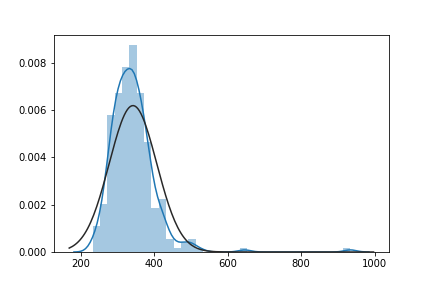}
    \label{focal_loss+MSE_loss}
    \end{minipage}%
    }
    \subfigure[Density Estimation of Inter-View Importance Latent Space via BOSS of Indoor User Movement Datasets(Between source view 1 and target view )]
    {
    \begin{minipage}{5cm}
    \centering
    \includegraphics[width=5cm]{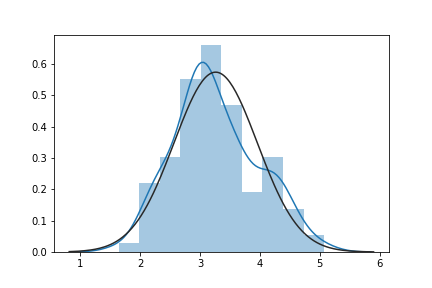}
    \label{focal_loss}
    \end{minipage}%
    }

      \centering
    \subfigure[Density Estimation of Inter-View Importance Latent Space via BOSS of Indoor User Movement Datasets(Between source view 2 and target view )]
    {
    \begin{minipage}{5cm}
    \centering
    \includegraphics[width=5cm]{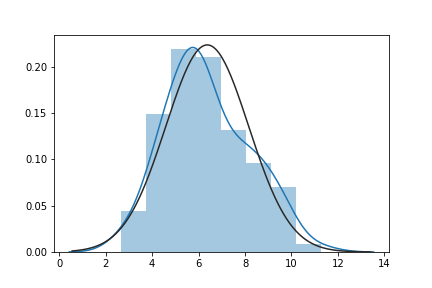}
    \label{focal_loss}
    \end{minipage}%
    }
    \subfigure[Density Estimation of Inter-View Importance Latent Space via BOSS of Indoor User Movement Datasets(Between source view 3 and target view )]
    {
    \begin{minipage}{5cm}
    \centering
    \includegraphics[width=5cm]{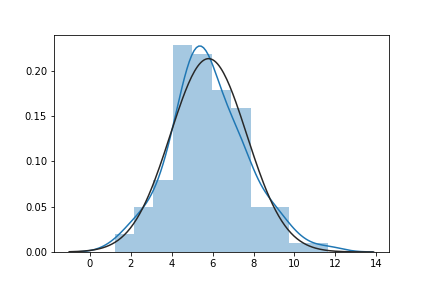}
    \label{focal_loss+MSE_loss}
    \end{minipage}%
    }
    \subfigure[Density Estimation of Inter-View Importance Latent Space via BOSS of Sports Activity Datasets(Between source view 1 and target view )]
    {
    \begin{minipage}{5cm}
    \centering
    \includegraphics[width=5cm]{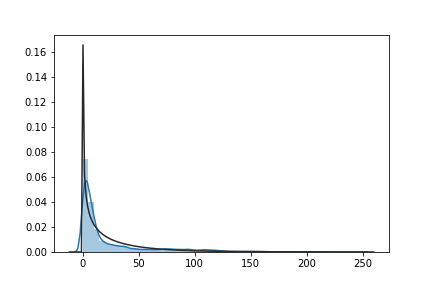}
    \label{focal_loss}
    \end{minipage}%
    }

    \centering
    \subfigure[Density Estimation of Inter-View Importance Latent Space via BOSS of Sports Activity Datasets(Between source view 2 and target view )]
    {
    \begin{minipage}{5cm}
    \centering
    \includegraphics[width=5cm]{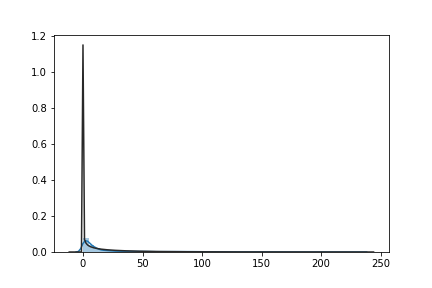}
    \label{focal_loss}
    \end{minipage}%
    }
    \subfigure[Density Estimation of Inter-View Importance Latent Space via BOSS of Sports Activity Datasets(Between source view 3 and target view )]
    {
    \begin{minipage}{5cm}
    \centering
    \includegraphics[width=5cm]{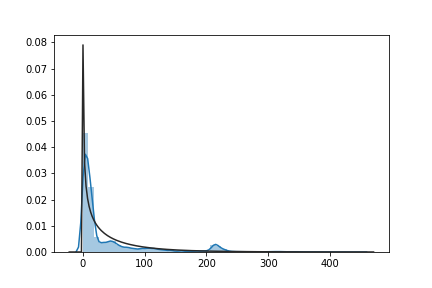}
    \label{focal_loss+MSE_loss}
    \end{minipage}%
    }
    \subfigure[Density Estimation of Inter-View Importance Latent Space via BOSS of Sports Activity Datasets(Between source view 4 and target view )]
    {
    \begin{minipage}{5cm}
    \centering
    \includegraphics[width=5cm]{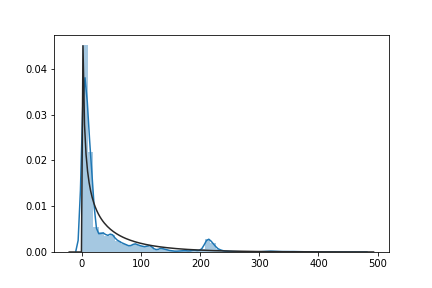}
    \label{focal_loss}
    \end{minipage}%
    }
    
\caption{Density Estimation in Latent Space for Different Datasets}
\label{Figure:dense}
\end{figure*}


\begin{thebibliography}{99}




  
\bibitem{y1}
P. D. Grunwald, {\em The Minimum Description Length Principle (Adaptive
Computation and Machine Learning)}, The MIT Press, 2007.

\bibitem{y2}
F. Petitjean, G. Forestier, G. I. Webb, A. E. Nicholson, Y. Chen,
and E. Keogh,{\em Dynamic Time Warping Averaging of Time Series
Allows Faster and More Accurate Classification},IEEE International
Conference on Data Mining, 2014, pp. 470–479.

\bibitem{y3}
H. I. Fawaz, G. Forestier, J. Weber, L. Idoumghar, and P. Muller,{\em Data
augmentation using synthetic data for time series classification with deep
residual networks}, CoRR, vol. abs/1808.02455, 2018.

\bibitem{y4}
S. Das Bhattacharjee, B. V. Balantrapu, W. Tolone, and A. Talukder,{\em Identifying extremism in social media with multi-view context-aware
subset optimization} , 2017 IEEE International Conference on Big
Data (Big Data), 2017, pp. 3638–3647.

\bibitem{y5}
M. Langkvist, L. Karlsson, and A. Loutfi,{\em A review of unsupervised
feature learning and deep learning for time-series modeling}, Pattern
Recognition Letters, pp. 11–24, 2014.

\bibitem{y6}
S. Li, Y. Li, and Y. Fu,{\em Multi-view time series classification: A
discriminative bilinear projection approach}, Proceedings of the
25th ACM International on Conference on Information and Knowledge
Management, ser. CIKM ’16, 2016, pp. 989–998.

\bibitem{y7}
Z. Cui, W. Chen, and Y. Chen,{\em Multi-Scale Convolutional Neural
Networks for Time Series Classification}, ArXiv, 2016.

\bibitem{y8}
I. Sutskever, O. Vinyals, and Q. V. Le, {\em Sequence to Sequence Learning
with Neural Networks}, Neural Information Processing Systems,
2014, pp. 3104–3112.

\bibitem{y9}
Y. Chen, E. Keogh, B. Hu, N. Begum, A. Bagnall, A. Mueen, and
G. Batista,{\em The UCR Time Series Classification Archive}, July 2015.

\bibitem{y10}
Z. Wang, W. Yan, and T. Oates, {\em Time series classification from
scratch with deep neural networks: A strong baseline}, CoRR, vol.
abs/1611.06455, 2016.

\bibitem{y11}
J. Cristian Borges Gamboa,{\em Deep Learning for Time-Series Analysis}, ArXiv, 2017.

\bibitem{y12}
S. Seto, W. Zhang, and Y. Zhou,{\em Multivariate time series classification
using dynamic time warping template selection for human activity recognition}, 2015 IEEE Symposium Series on Computational Intelligence, pp.
1399–1406, 2015.

\bibitem{y13}
P.-F. Marteau and S. Gibet,{\em On recursive edit distance kernels with
application to time series classification},IEEE transactions on neural
networks and learning systems, vol. 26, no. 6, June 2015.

\bibitem{y14}
J. Lines and A. Bagnall,{\em Time series classification with ensembles of
elastic distance measures}, Data Min. Knowl. Discov., vol. 29, no. 3, pp.
565–592, May 2015.

\bibitem{y15}
E. Keogh and S. Kasetty,{\em On the need for time series data
mining benchmarks: a survey and empirical demonstration},Data Mining and Knowledge Discovery, 7(4):349–371, 2003.

\bibitem{y16}
Z. Xing, J. Pei, and S. Y. Philip, {\em Early classification on
time series}, Knowledge and information systems,
31(1):105–127, 2012.

\bibitem{y17}
A. Blum and T. Mitchell, {\em Combining labeled and unlabeled
data with co-training}, In Proceedings of the Eleventh
Annual Conference on Computational Learning Theory,
pages 92–100. ACM, 1998.

\bibitem{y18}
C. Xu, D. Tao, and C. Xu, {\em A survey on multi-view
learning}, arXiv preprint arXiv:1304.5634, 2013.

\bibitem{y19}
Z. Fang and Z. Zhang, {\em Simultaneously combining
multi-view multi-label learning with maximum margin
classification}, In Proceedings of IEEE International
Conference on Data Mining, pages 864–869. IEEE, 2012.

\bibitem{y20}
J. Yosinski, J. Clune, Y. Bengio, and H. Lipson,{\em How transferable are
features in deep neural networks?}, in Advances in Neural Information
Processing Systems 27, Z. Ghahramani, M. Welling, C. Cortes, N. D.
Lawrence, and K. Q. Weinberger, Eds., 2014.

\bibitem{y21}
G. Csurka,{\em Domain adaptation for visual applications: A comprehensive
survey}, CoRR, vol. abs/1702.05374, 2017.

\bibitem{y22}
A. T. Sreyasee Das Bhattacharjee,{\em “Graph clustering for weapon
discharge event detection and tracking in infrared imagery using deep
features}, 2017. [Online]. Available: https://doi.org/10.1117/12.2277737

\bibitem{y23}
S. Das Bhattacharjee, B. V. Balantrapu, W. Tolone, and A. Talukder,{\em Identifying extremism in social media with multi-view context-aware
subset optimization}, 2017 IEEE International Conference on Big
Data (Big Data), 2017, pp. 3638–3647.

\bibitem{y24}
S. Das Bhattacharjee, A. Talukder, and B. V. Balantrapu,{\em Active
learning based news veracity detection with feature weighting and deepshallow
fusion}, 2017 IEEE International Conference on Big Data
(Big Data), 2017, pp. 556–565.

\bibitem{y25}
S. Das Bhattacharjee, V. S. Paranjpe, and W. Tolone, {\em Identifying
malicious social media contents using multi-view context-aware active
learning}, Future Generation Computer Systems, Elsevier, 2017.

\bibitem{y26}
S. Das Bhattacharjee, J. Yuan, Z. Jiaqi, and Y. Tan,{\em Context-aware
graph-based analysis for detecting anomalous activities}, 2017 IEEE
International Conference on Multimedia and Expo (ICME), 2017, pp.
1021–1026.

\bibitem{y27}
Y. Geng and X. Luo, {\em Cost-Sensitive Convolution based Neural Networks for Imbalanced Time-Series Classification}, ArXiv e-prints, 2018.

\bibitem{y28}
A. Ziat, E. Delasalles, L. Denoyer, and P. Gallinari, {\em Spatio-Temporal Neural Networks for Space-Time Series Forecasting and Relations Discovery},
IEEE International Conference on Data Mining, 2017, pp. 705–714.

\bibitem{y29}
Z. Che, Y. Cheng, S. Zhai, Z. Sun, and Y. Liu,{\em Boosting Deep Learning Risk Prediction with Generative Adversarial Networks for Electronic Health Records}, IEEE International Conference on Data Mining, 2017, pp. 787–792.

\bibitem{y30}
H. Ismail Fawaz, G. Forestier, J. Weber, L. Idoumghar, and P.-A. Muller,{\em Deep learning for time series classification: a review}, ArXiv, 2018.

\bibitem{y31}
M. Baktashmotlagh, M. Faraki, T. Drummond, and M. Salzmann,{\em Learning factorized representations for open-set domain adaptation}, CoRR, vol. abs/1805.12277, 2018.

\bibitem{y32}
M. Long and J. Wang, {\em Learning transferable features with deep
adaptation networks}, CoRR, vol. abs/1502.02791, 2015.

\bibitem{y33}
V. Vercruyssen, W. Meert, and J. Davis,{\em Transfer Learning for Time Series Anomaly Detection}, Workshop and Tutorial on Interactive Adaptive Learning co-located with European Conference on Machine Learning and Principles and Practice of Knowledge Discovery in Databases, 2017, pp. 27–36.

\bibitem{y34}
D. Zhan, S. Yi and D. Jiang, {\em Small-Scale Demographic Sequences Projection Based on Time Series Clustering and LSTM-RNN.} ICDM Workshops 2018.

\bibitem{y35}
J. Serra, S. Pascual, and A. Karatzoglou, {\em Towards a universal neural network encoder for time series},CoRR, vol. abs/1805.03908, 2018.

\bibitem{y36}
Y. Guo, {\em Convex subspace representation learning from multi-view data},
 Proceedings of the 27th AAAI Conference on Artificial Intelligence, volume 1, page 2, 2013.
 
\bibitem{y37}
Y. Li, F. Nie, H. Huang, and J. Huang,{\em Large-scale multi-view spectral clustering via bipartite graph}, Proceedings of the Twenty-Eighth AAAI Conference on Artificial Intelligence, pages 2750–2756, 2015.

\bibitem{y38}
W. Wang, R. Arora, K. Livescu, and J. Bilmes, {\em On deep multi-view representation learning},  Proceedings of the 32nd International Conference on Machine Learning, pages 1083–1092, 2015.

\bibitem{y39}
M. Kan, S. Shan, H. Zhang, S. Lao, and X. Chen, {\em Multi-view discriminant analysis}, IEEE Transactions on Pattern Analysis and Machine Intelligence, 38(1):188–194, 2016.

\bibitem{y40}
P.-Y. Zhou and K. C. Chan, {\em A feature extraction method for multivariate time series classification using temporal patterns}, Advances in Knowledge Discovery and Data
Mining, pages 409–421. Springer, 2015.

\bibitem{y41}
H. Hayashi, T. Shibanoki, K. Shima, Y. Kurita, and T. Tsuji,{\em  A recurrent probabilistic neural network with dimensionality reduction based on time-series discriminant component analysis}, IEEE Transactions on Neural Networks and Learning Systems, 26(12):3021–3033, 2015.

\bibitem{y42}
Y. Zheng, Q. Liu, E. Chen, J. L. Zhao, L. He, and G. Lv, {\em Convolutional nonlinear neighbourhood components analysis for time series classification}, Advances in Knowledge Discovery and Data Mining, pages 534–546. Springer, 2015.

\bibitem{y43}

S. Yi, D. Zhan, Z. Geng, W. Zhang and C. Xu, {\em FIS-GAN: GAN with Flow-based Importance Sampling}, arXiv, preprint, arXiv:1910.02519.

\bibitem{y44}
M. Lichman. UCI machine learning repository, 2013 .


\end{thebibliography}
\end{document}